# CPMLHO:Hyperparameter Tuning via Cutting Plane and Mixed-Level Optimization

**Shuo Yang[1*], Yang Jiao[1], Shaoyu Dou[1], Mana Zheng[1], Chen Zhu[1]**

[1]Department of Computer Science and Technology, Tongji University, Shanghai, 201804, China

*Corresponding author e-mail: 2033086@tongji.edu.cn

**Abstract.** The hyperparameter optimization of neural network can be expressed as a bilevel optimization problem. The bilevel optimization is used to automatically update the hyperparameter, and the gradient of the hyperparameter is the approximate gradient based on the best response function. Finding the best response function is very time consuming. In this paper we propose CPMLHO, a new hyperparameter optimization method using cutting plane method and mixed-level objective function. The cutting plane is added to the inner layer to constrain the space of the response function. To obtain more accurate hypergradient, the mixed-level can flexibly adjust the loss function by using the loss of the training set and the verification set. Compared to existing methods, the experimental results show that our method can automatically update the hyperparameters in the training process, and can find more superior hyperparameters with higher accuracy and faster convergence.

## 1. Introduction

With the rapid development of deep learning, it is essential to adjust hyperparameters in order to achieve the optimal performance in challenging datasets such as image datasets[6]. Traditional hyperparameter optimization methods, such as grid search, random search[2] and Bayesian optimization[20], perform well in the low-dimensional hyperparameter spaces, but these methods need to fix the initial value of the hyperparameter, require many training runs and a lot of computing resources.

The gradient-based bilevel optimization method has achieved excellent performance in hyperparameter optimization[7], meta-learning[19] and neural network search[5]. Formally, the hyperparameter gradient-based method can be expressed as a bilevel optimization problem[12]:

$$\boldsymbol{\lambda}^* = \arg\min_{\boldsymbol{\lambda}} L_{val}(\mathbf{w}^*, \boldsymbol{\lambda})$$

$$s.t. \quad \mathbf{w}^* = \arg\min_{\mathbf{w}} L_{tr}(\mathbf{w}, \boldsymbol{\lambda}) \tag{1}$$

$$var \quad \mathbf{w} \in \mathbb{R}^m, \boldsymbol{\lambda} \in \mathbb{R}^n.$$

$\mathbf{w} \in \mathbb{R}^m$ represents network weight, $\boldsymbol{\lambda} \in \mathbb{R}^n$ determines the hyperparameters, $L_{tr}$ and $L_{val}$ denote the losses with respect to training data and validation data with $\mathbf{w}$ and $\boldsymbol{\lambda}$ respectively.

At present, the hyperparameter optimization in bilevel optimization is mainly based on implicit function[16], dynamical system[18] and hypernetwork[13]. The implicit differentiation method needs to calculate the inverse Hessian. The dynamical system based method needs the inverse or forward gradient calculation. These two methods are computationally expensive. The hypernetwork method uses the best response function $\mathbf{w}(\boldsymbol{\lambda}) = \arg\min_{\mathbf{w}} L_{tr}(\mathbf{w}, \boldsymbol{\lambda})$ to jointly update the hypernetwork and the hyperparameters. In order to find the best response function in the inner layer, self-tuning network(STN)[15] constructs a hypernetwork for each layer. $\Delta$-STN[1] uses the the best-response Jacobian to get the optimal solution. It is difficult to find the best response function for bilevel optimization based on hypernetwork. The time cost is very high, which requires hundreds of training epochs. And the deviation of the best response function will lead to the deviation of the approximate gradient and the true gradient of the hyperparameter.

To solve the problem that the best response function is difficult to find, we introduce the cutting plane method to approximate the feasible region of an optimization problem[22,23]. Cutting plane method is mainly used in solving integer linear programming[8,24] and convex optimization problems[14].It can also be used to tackle non-convex optimization problems[17]. We use the cutting plane method to limit the search space of the response function and find the optimal solution more quickly. In addition, cutting plane method also facilitates distributed implementation[21].

In addition, the bilevel optimization only uses the verification set loss to calculate the hyperparameter gradient to update the hyperparameter. This does not utilize the potential relationship between the training set and the verification set, and uses the approximate gradient based on the approximation of $\mathbf{w}$, causing gradient error. Mixed level was originally used in neural network architecture search[9]. We introduce mixed level objective function to flexibly adjust the loss of the training set and the verification set when calculating the gradient of the hyperparameter.

In this work, we use the bilevel optimization method based on the hypernetwork to update the hyparameters. The gradient of the inner layer is calculated to update the network parameters and response function, and the gradient of the outer layer is calculated to update the hyperparameter. When searching for the best response function in the inner layer, we use the cutting plane method to constrain the search scope of the response function until the best response function is searched. Finally, we use mixed level objective function to avoid the gradient error caused by the loss of only using the verification set when calculating the gradient of the hyperparameter in the outer layer. The gradient of the training set and the verification set guarantees the updating direction of the hyperparameter gradient, which can make the hyperparameters more stable and accurate. Experiments show that compared with existing method, our method can find more stable and accurate hyperparameters and achieve superior performance in different datasets.

2. Related work
Colson provides the idea of a bilevel problem[4]. Snell used bilevel optimization program to optimize meta-learning to learn the common parameters of all tasks[19]. Cortes et al. describe neural network architecture search as a bilevel problem, and use bilevel optimization to find the best neural network architecture[5]. Using bilevel optimization to adjust hyperparameters not only uses rich gradient information, but also makes full use of network structure and expands the optimization of the number of hyperparameters. Recently, a variety of bilevel optimization methods for hyperparameter optimization have been proposed, including implicit differentiation based method[16], dynamic systems based method[18] and hypernetwork based method[13]. Methods based on implicit differentiation and dynamic systems introduce a large computational overhead. Based on hypernetwork, such as self-tuning network (STN)[15], the best response function is used to calculate the hypergradient of the hyperparameter and automatically update the hyperparameter. $\Delta$-STN[1] improves some of the shortcomings of STN by using a structured response jacobians. The hypernetwork based method is very dependent on finding the best response function. Once the deviation occurs, the approximate gradient of the outer function to the hyperparameter will be wrong. And the search for the best response function also brings a lot of time overhead.

The cutting plane method is mainly used in integer linear programming such as Gomory's cut[8], Chvatal -- Gomory cut[3], implied bound cut[11]. Huang used the cutting plane method to solve the neural network verification problem[23]. Cutting plane method is an effective convex optimization method. Convex functions are subdifferentiable anywhere in their domain[10]. Cutting plane method can solve all convex problems, including those where the unavailable gradient applies to both differentiable and non-differentiable[14]. We use cutting plane method to constrain the search space of the response function, so as to find the best response function.

**3. Method**

*3.1. Problem statement*
Our optimization problem can be expressed as:
$$\boldsymbol{\lambda}^* = \arg\min_{\boldsymbol{\lambda}} L(\mathbf{w}^*(\boldsymbol{\lambda}), \boldsymbol{\lambda})$$
$$s.t. \quad \mathbf{w}^*(\boldsymbol{\lambda}) = \arg\min_{\mathbf{w}(\boldsymbol{\lambda})} l(\mathbf{w}(\boldsymbol{\lambda}), \boldsymbol{\lambda}) \tag{2}$$
$$var \quad \mathbf{w}(\boldsymbol{\lambda}) \in \mathbb{R}^m, \boldsymbol{\lambda} \in \mathbb{R}^n.$$

$\boldsymbol{\lambda}$ is the hyperparameters. $\mathbf{w}$ is the weight of the model. $L$ Is the outer loss function, $l$ is the inner loss function. When the inner layer function is optimal, the outer layer optimization is carried out to obtain the hyperparameter gradient and achieve the purpose of optimizing the hyperparameter.

*3.2. The gradient of the hyperparameter*
$\mathbf{w}(\boldsymbol{\lambda})$ is the response function of the parameter to the hyperparameter:
$$\mathbf{w}(\boldsymbol{\lambda}) = \mathbf{w}_e + \mathbf{w}_h = \mathbf{w}_e + (\mathbf{w}_{h1}\boldsymbol{\lambda})\mathbf{w}_{h2}. \tag{3}$$
$\mathbf{w}_e$ is the weight of the neural network from input to output of the model. $\mathbf{w}_h$ consists of two parts, $\mathbf{w}_{h1}$ is the mapping of hyperparameters to network weights, and $\mathbf{w}_{h2}$ is the network weights of the model from input to output.
The output predicted $\hat{y}$ by the model is:
$$\hat{y}(x; \mathbf{w}(\boldsymbol{\lambda})) = \mathbf{w}(\boldsymbol{\lambda})x = \mathbf{w}_e x + (\mathbf{w}_{h1}\boldsymbol{\lambda})\mathbf{w}_{h2} x. \tag{4}$$
The inner loss function of the model is:
$$l(\boldsymbol{\lambda}, \mathbf{w}(\boldsymbol{\lambda})) = \sum_{(x,y) \in D} (\hat{y} - y)^2. \tag{5}$$
Finally, the gradient of the hyperparameter in the outer layer is:
$$\nabla_{\boldsymbol{\lambda}} = \nabla_{\boldsymbol{\lambda}} L(\mathbf{w}(\boldsymbol{\lambda}), \boldsymbol{\lambda}) + \nabla_{\mathbf{w}(\boldsymbol{\lambda})} L(\mathbf{w}(\boldsymbol{\lambda}), \boldsymbol{\lambda}) * \nabla_{\boldsymbol{\lambda}} \mathbf{w}(\boldsymbol{\lambda}). \tag{6}$$

*3.3. Add the cutting plane to constrain the search scope of the response function:*
From Equation 6, we can see that the gradient of the hyperparameter depends not only on the direct gradient but also on the response function $\mathbf{w}(\boldsymbol{\lambda})$. Getting the best response function $\mathbf{w}^*(\boldsymbol{\lambda})$ is very difficult and time consuming. Using cutting plane method, we can limit the search space of $\mathbf{w}(\boldsymbol{\lambda})$ and approximate the optimal solution.
The inner optimization objective function is equivalent to:
$$\min \; l(\mathbf{w}_e, \mathbf{w}_h) = \sum_{(x,y) \, D} (\hat{y} - y)^2 = \sum_{(x,y) \, D} (\mathbf{w}_e x + \mathbf{w}_h x - y)^2. \tag{7}$$
Inner optimization problem can be viewed as a constrained problem:
$$\mathbf{w}_h = \arg\min_{\mathbf{w}_h} l(\mathbf{w}_e, \mathbf{w}_h) \tag{8}$$
$$s.t. \quad \mathbf{w}_e = \arg\min_{\mathbf{w}_e} l(\mathbf{w}_e, \mathbf{w}_h). \tag{9}$$
We define $\mathbf{g}(\mathbf{w}_e, \mathbf{w}_h) = |\mathbf{w}_e - \mathbf{w}_h|$. After convex relaxation, the inner optimization is equivalent to:

$$\min\ l(\mathbf{w}_e, \mathbf{w}_h) \quad where\ \{\mathbf{g}(\mathbf{w}_e, \mathbf{w}_h) \leq \varepsilon|\}. \tag{10}$$

After k steps of iteration in the inner layer, we add the cutting plane constraint:

$$\mathbf{g}^k + \frac{\mathbf{g}^k}{\mathbf{w}_e^k}(\mathbf{w}_e - \mathbf{w}_e^k) + \frac{\mathbf{g}^k}{\mathbf{w}_h^k}(\mathbf{w}_h - \mathbf{w}_h^k) \leq \varepsilon^2. \tag{11}$$

$\varepsilon^2$ is infinitesimal, $\frac{\mathbf{g}^k}{\mathbf{w}_e^k}$ is the derivative of $\mathbf{g}^k$ with respect to $\mathbf{w}_e^k$, $\frac{\mathbf{g}^k}{\mathbf{w}_h^k}$ is the derivative of $\mathbf{g}^k$ with respect to $\mathbf{w}_h^k$. In the process of training $\mathbf{w}_h$, there is not only the weight $\mathbf{w}_{h2}$ from input to output, but also the mapping of $\mathbf{w}_{h1}$ from $\boldsymbol{\lambda}$ to $\mathbf{w}_{h1}$. However, $\mathbf{w}_{h1}$ does not contain input information, which makes $\mathbf{w}_h$ increase interference in the process of gradient descent, so $\mathbf{w}_h$ is not as accurate as $\mathbf{w}_e$. Therefore, the reason why we add this constraint is that $\mathbf{w}_h$ could approximate $\mathbf{w}_e$ as much as possible to obtain a more accurate response function $\mathbf{w}(\boldsymbol{\lambda})$.

We add additional constraints to the inner loss function when we update $\mathbf{w}_h$:

$$\varphi_{\mathbf{w}_e,\mathbf{w}_h} = \mathbf{g}^k + \frac{\mathbf{g}^k}{\mathbf{w}_e^k}(\mathbf{w}_e - \mathbf{w}_e^k) + \frac{\mathbf{g}^k}{\mathbf{w}_h^k}(\mathbf{w}_h - \mathbf{w}_h^k) - \varepsilon^2. \tag{11}$$

$$l(\mathbf{w}_e, \mathbf{w}_h, \{\mu, \varphi\}) = l(\mathbf{w}_e, \mathbf{w}_h) + \mu * \varphi_{\mathbf{w}_e,\mathbf{w}_h}. \tag{12}$$

$\mu$ is the lagrange multiplier. The meaning of Equation 12 is to solve the minimum value of the inner objective function $l(\mathbf{w}_e, \mathbf{w}_h)$ under constraints $\varphi_{\mathbf{w}_e,\mathbf{w}_h}$. By using the gradient descent algorithm we update $\mathbf{w}_e$ with no additional constraints and $\mathbf{w}_h$ with constraints respectively:

$$\mathbf{w_e} = \mathbf{w_e} - \alpha_{\mathbf{w}_e} \nabla_{\mathbf{w}_e} l(\mathbf{w}_e, \mathbf{w}_h) \tag{13}$$

$$\mathbf{w_h} = \mathbf{w_h} - \alpha_{\mathbf{w}_h} \nabla_{\mathbf{w}_h} (l(\mathbf{w}_e, \mathbf{w}_h) + \mu * \varphi_{\mathbf{w}_e,\mathbf{w}_h}). \tag{14}$$

*3.4. Use mixed-level objective function to update the hyperparameter:*

When calculating the hyperparameter gradient in the outer layer, the previous bilevel optimization method only uses the loss of the verification set, without considering the potential relationship between the training set and the verification set. We use the mixed-level objective function to calculate the gradient of the hyperparameters by the loss of training set and the loss of verification set, and make full use of the information of dataset to get better hyperparameters. The mixed-level objective function can be expressed as:

$$\min_{\boldsymbol{\lambda}} L_{tr}(\mathbf{w}^*(\boldsymbol{\lambda}), \boldsymbol{\lambda}) + \theta * L_{val}(\mathbf{w}^*(\boldsymbol{\lambda}), \boldsymbol{\lambda}). \tag{15}$$

$\theta$ is a non-negative regularization parameter that balances the importance of the training loss and validation loss.

The gradient of the hyperparameter is:

$$\min_{\boldsymbol{\lambda}} L_{tr}(\mathbf{w}^*(\boldsymbol{\lambda}), \boldsymbol{\lambda}) + \theta * L_{val}(\mathbf{w}^*(\boldsymbol{\lambda}), \boldsymbol{\lambda}). \tag{16}$$

When $\theta$ is 0, we can look at it as a single level optimization, and when $\theta$ is infinity, we can look at it as a bilevel optimization. When updating the hyperparameters, the training loss can effectively judge the performance of the neural network, and the validation set loss can also judge the underfitting or overfitting state of the neural network. By combining the gradient of the two, the gradient direction of the hyperparameters can be effectively controlled to find more effective hyperparameters.

## 4. Experiments

In this section, we conducted experiments on image classification datasets Mnist, FashionMnist, all on NVIDIA MX350. We compared our methode to random search (RS) ,Bayesian optimization (BO) , and $\Delta$-STN.The final experimental performance on validation loss are shown in Table 1.

Table 1. The final validation loss of each method on the Mnist and FashionMnist

|  | network | RS | BO | $\Delta$-STN | CPMLHO |
|---|---|---|---|---|---|
| Mnist | MLP | 0.183 | 0.182 | 0.182 | **0.180** |
| FashionMnist | CNN | 0.548 | 0.541 | 0.531 | **0.518** |

### 4.1. Mnist

For Mnist dataset, we trained a multi-layer perceptron consisting of three linear layers.We adjusted the dropout of each layer. We added cutting plane polynomial constraints after each training epoch.The training curves of each layer of dropout are shown in Figure 1.From the training curve, we can see that our method has improved to the most suitable for the model's hyperparameters at the iteration of 200 steps, while the hyperparameter update curve of the $\Delta$-STN method is relatively slow, and even changes the iteration direction of dropout0. This indicates that our method can find the hyperparameter more quickly and is more stable than $\Delta$-STN in the updating process.

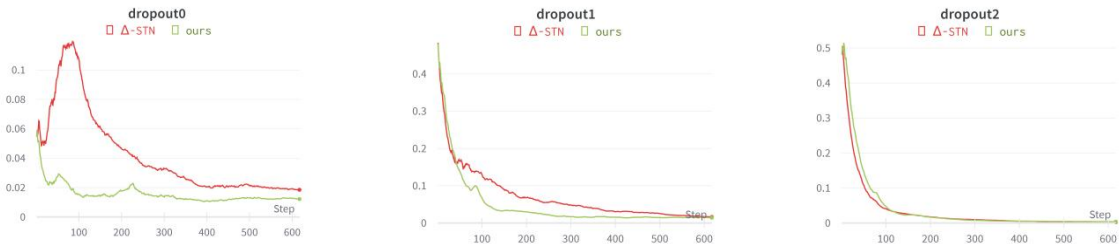

Figure 1. The hyperparameter schedule prescribed by $\Delta$-STN and our method

### 4.2. FashionMnist

We trained the FashionMnist dataset on a neural network consisting of 2 layers of CNN and a linear layer. We adjusted 5 hyperparameters: the dropout of each layer,the cutout holes and length. The final values of the hyperparameters are shown in Table 2.

Table 2. The final values of the hyperparameters of each method on the FashionMnist

|  | Dropout0 | Dropout1 | Dropout2 | Cutout holes | Cutout length |
|---|---|---|---|---|---|
| $\Delta$-STN | 0.008 | 0.062 | 0.044 | 0.86 | 2.2 |
| CPMLHO | 0.019 | 0.149 | 0.057 | 0.73 | 3.3 |

Table 1,3 show that under the hyperparameters we found, the model has better performance in validation loss, validation accuracy and test accuracy than the previous traditional hyperparameter method and STN method.

Table 3. Final performance of each method on the image classification tasks

|  | Mnist | | FashionMnist | |
|---|---|---|---|---|
| Method | Val Acc | Test acc | Val Acc | Test acc |
| $\Delta$-STN | 0.946 | 0.941 | 0.809 | 0.798 |
| CPMLHO | **0.947** | **0.942** | **0.815** | **0.809** |

*4.3. Hyperparametric sensitivity analysis*

We also did another experiment on the Mnist dataset. We set different initial values of 0.1, 0.2, 0.5 and 0.8 for dropout0. The training curves of dropout0 are shown in Figure 2. The dropout0 will eventually drop to around 0.05, which indicates that our method is not sensitive to the initial value set and can accurately find the direction of the gradient of the hyperparameter.

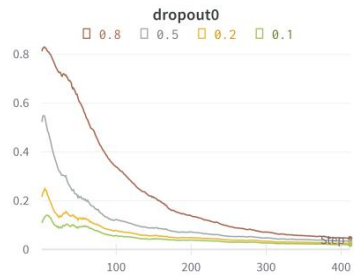

**Figure 2.** The final value of dropout0 for different initial dropout0

## 5. Conclusion

We map the hyperparameters to the network weight by approximating the optimal response function, and use the gradient descent method to automatically update the hyperparameters. At the same time, we use the cutting plane method to constrain the search space of the response function, thus obtaining a response function with better performance. Finally, we use mixed level optimization to update the hyperparameters by mixing the loss of training set and verification set, which can reduce the approximate gradient error brought by the bilevel optimization. Experiments show that our method can not only determine the updating direction of the hyperparameters faster, but also search for better hyperparameters, which brings improvement to the experimental results. We believe that our method provides a more reliable, more efficient and more accurate hyperparameter optimization method, and provides a useful for neural network hyperparameter tuning.

## 6. References

[1] Juhan Bae and Roger B Grosse. Delta-stn: Efficient bilevel optimization for neural networks using structured response jacobians. Advances in Neural Information Processing Systems,33:21725–21737, 2020.
[2] James Bergstra and Yoshua Bengio. Random search for hyperparameter optimization. Journal of machine learning research,13(2), 2012.
[3] Vasek Chvátal. Edmonds polytopes and a hierarchy of combinatorial problems. Discrete mathematics, 4(4):305–337,1973.
[4] Benoît Colson, Patrice Marcotte, and Gilles Savard. An overview of bilevel optimization. Annals of operations research,153(1):235–256, 2007.
[5] Corinna Cortes, Xavier Gonzalvo, Vitaly Kuznetsov, Mehryar Mohri, and Scott Yang. Adanet: Adaptive structural learning of artificial neural networks. In International conference on machine learning, pages 874–883. PMLR,2017.
[6] Jia Deng, Wei Dong, Richard Socher, Li-Jia Li, Kai Li, and Li Fei-Fei. Imagenet: A large-scale hierarchical image database.In 2009 IEEE conference on computer vision and pattern recognition, pages 248–255. Ieee,2009.

[7]     Matthias Feurer and Frank Hutter. Hyperparameter optimization. In Automated machine learning, pages 3–33. Springer,Cham, 2019.

[8]     Paul C Gilmore and Ralph E Gomory. A linear programming approach to the cutting-stock problem. Operations research,9(6):849–859, 1961.

[9]     Chaoyang He, Haishan Ye, Li Shen, and Tong Zhang. Milenas:Efficient neural architecture search via mixed-level reformulation. In Proceedings of the IEEE/CVF Conference on Computer Vision and Pattern Recognition, pages 11993–12002,2020.

[10]    Jean-Baptiste Hiriart-Urruty and Claude Lemaréchal. Convex analysis and minimization algorithms I: Fundamentals, volume 305. Springer science & business media,2013.

[11]    Karla L Hoffman and Manfred Padberg. Solving airline crew scheduling problems by branch-and-cut. Management science,39(6):657–682, 1993.

[12]    Hanxiao Liu, Karen Simonyan, and Yiming Yang. Darts: Differentiable architecture search. arXiv preprint arXiv:1806.09055,2018.

[13]    Jonathan Lorraine and David Duvenaud. Stochastic hyperparameter optimization through hypernetworks. arXiv preprint arXiv:1802.09419, 2018.

[14]    Meng-long Lu, Lin-bo Qiao, Da-wei Feng, Dong-sheng Li, and Xi-cheng Lu. Mini-batch cutting plane method for regularized risk minimization. Frontiers of Information Technology & Electronic Engineering, 20(11):1551–1563,2019.

[15]    Matthew MacKay, Paul Vicol, Jon Lorraine, David Duvenaud, and Roger Grosse. Self-tuning networks: Bilevel optimization of hyperparameters using structured best-response functions.arXiv preprint arXiv:1903.03088,2019.

[16]    Fabian Pedregosa. Hyperparameter optimization with approximate gradient. In International conference on machine learning,pages 737–746. PMLR, 2016.

[17]    Zhi Pei, Mingzhong Wan, Zhong-Zhong Jiang, Ziteng Wang,and Xu Dai. An approximation algorithm for unrelated parallel machine scheduling under tou electricity tariffs. IEEE Transactions on Automation Science and Engineering, 18(2):743–756,2021

[18]    Amirreza Shaban, Ching-An Cheng, Nathan Hatch, and Byron Boots. Truncated back-propagation for bilevel optimization. In The 22nd International Conference on Artificial Intelligence and Statistics, pages 1723–1732. PMLR,2019.

[19]    Jake Snell, Kevin Swersky, and Richard Zemel. Prototypical networks for few-shot learning. Advances in neural information processing systems, 30,2017.

[20]    Jasper Snoek, Hugo Larochelle, and Ryan P Adams. Practical bayesian optimization of machine learning algorithms. Advances in neural information processing systems, 25,2012.

[21]    Jiao Yang, Yang Kai, and Dongjing Song. Distributed distributionally robust optimization with non-convex objectives.Advances in neural information processing systems,35,2022.

[22]    K. Yang, Y. Wu, J. Huang, X. Wang, and S. Verdu. Distributed robust optimization for communication networks. In IEEE INFOCOM 2008 - The 27th Conference on Computer Communications, pages 1157–1165,2008.

[23]    Kai Yang, Jianwei Huang, Yihong Wu, Xiaodong Wang, and Mung Chiang. Distributed robust optimization (DRO), part I: Framework and example. Optimization and Engineering,15(1):35–67, 2014.

[24]    Kai Yang, Xiaodong Wang, and Jon Feldman. A new linear programming approach to decoding linear block codes. IEEE Transactions on Information Theory,54(3):1061-1072,2008..

[25]    Huan Zhang, Shiqi Wang, Kaidi Xu, Linyi Li, Bo Li, Suman Jana, Cho-Jui Hsieh, and J Zico Kolter. General cutting planes for bound-propagation-based neural network verification. arXiv preprint arXiv:2208.05740, 2022.